\documentclass[10pt,twocolumn,letterpaper]{article}

\usepackage[pagenumbers]{wacv} % To force page numbers, e.g. for an arXiv version

\usepackage{graphicx}
\usepackage{amsmath}
\usepackage{amssymb}
\usepackage{booktabs}
\usepackage{gensymb}
\usepackage{xcolor}

\usepackage[pagebackref,breaklinks,colorlinks]{hyperref}

\usepackage[capitalize]{cleveref}
\crefname{section}{Sec.}{Secs.}
\Crefname{section}{Section}{Sections}
\Crefname{table}{Table}{Tables}
\crefname{table}{Tab.}{Tabs.}

\def\wacvPaperID{641} % *** Enter the WACV Paper ID here
\def\confName{WACV}
\def\confYear{2024}

\begin{document}

%\title{Demographic and Non-Demographic Diversity and Bias in Synthetic Face Recognition}
\title{Bias and Diversity in Synthetic-based Face Recognition}

\author{Marco Huber$^{1,2}$, Anh Thi Luu$^{1}$, Fadi Boutros$^{1}$, Arjan Kuijper$^{1,2}$, Naser Damer$^{1,2}$\\
$^{1}$ Fraunhofer Institute for Computer Graphics Research IGD, Darmstadt, Germany\\
$^{2}$ Department of Computer Science, TU Darmstadt,
Darmstadt, Germany\\
Email: marco.huber@igd.fraunhofer.de}

\maketitle

%%%%%%%%% ABSTRACT
\begin{abstract}
Synthetic data is emerging as a substitute for authentic data to solve ethical and legal challenges in handling authentic face data. The current models can create real-looking face images of people who do not exist. However, it is a known and sensitive problem that face recognition systems are susceptible to bias, i.e. performance differences between different demographic and non-demographics attributes, which can lead to unfair decisions. In this work, we investigate how the diversity of synthetic face recognition datasets compares to authentic datasets, and how the distribution of the training data of the generative models affects the distribution of the synthetic data. To do this, we looked at the distribution of gender, ethnicity, age, and head position. Furthermore, we investigated the concrete bias of three recent synthetic-based face recognition models on the studied attributes in comparison to a baseline model trained on authentic data. Our results show that the generator generate a similar distribution as the used training data in terms of the different attributes. With regard to bias, it can be seen that the synthetic-based models share a similar bias behavior with the authentic-based models. However, with the uncovered lower intra-identity attribute consistency seems to be beneficial in reducing bias.
\end{abstract}

%%%%%%%%% BODY TEXT
\vspace{-5mm}
\section{Introduction}
\vspace{-2mm}
% Establish Importance
Recently, training face recognition (FR) models on synthetic data and using FR models trained on synthetic data has gained attention and importance \cite{DBLP:journals/corr/abs-2305-01021}. {\color{black} Motivated by legal and ethical issues in some jurisdictions (e.g. the European Union \cite{eu-2016/679}) regarding using and sharing authentic face images without consent, synthetic data might be a substitute due to its remarkable quality and similarity to authentic data. Additionally, several large face datasets such as the widely-used MS-Celeb-1M \cite{DBLP:conf/eccv/GuoZHHG16}, VGGFace2 \cite{DBLP:conf/fgr/CaoSXPZ18} or DukeMTMC-reID \cite{DBLP:conf/eccv/RistaniSZCT16} are not available anymore from an official source.} However, the existing accuracy gap between authentic and synthetic-based FR requires novel solutions to train suitable well-performing FR models \cite{DBLP:conf/icb/BoutrosHSRD22, DBLP:conf/fgr/BoutrosKFKD23}.

In authentic  FR systems and datasets, the presence of bias, unfair behavior by the systems based on demographic (e.g. gender, age, ethnicity) or non-demographic (e.g. head-pose, accessories) attributes, has been studied extensively in the past \cite{9534882, DBLP:conf/eccv/AlviZN18, DBLP:journals/tbbis/CavazosPCO21} and mitigation strategies to increase data diversity and reduce unfair behavior have been explored \cite{DBLP:conf/eccv/AlviZN18, DBLP:conf/eccv/GongLJ20, DBLP:journals/prl/TerhorstKDKK20}. Having fairer models is of high importance for biometric applications as the decisions of FR systems are often of high impact on the individuals' lives, e.g. in access control or law enforcement identification.

For synthetic FR data and models trained on synthetic data, unfair behavior is still an unexplored area \cite{DBLP:journals/corr/abs-2305-01021}. The diversity and bias in synthetic data and models trained on synthetic data are of special interest as it provides novel opportunities but also new problems. A major advantage of synthetic data is that new data can be generated depending on certain attributes. For example, a lack of variety with respect to a certain ethnicity can be compensated by synthetic images of these underrepresented ethnicities \cite{DBLP:journals/corr/abs-2304-07060}. {\color{black} While there are works \cite{DBLP:conf/wacv/BhattaABK23, DBLP:conf/icb/AlbieroZB20} that have shown that the distribution of training data plays an exclusive role for gender bias, there are works \cite{DBLP:journals/pami/WangZD22, DBLP:conf/iccv/WangDHTH19} that showed that the distribution of training data plays a major role in ethnicity bias.} However, a drawback of synthetic data is, that there is no {\color{black} factual or} self-reported ground truth regarding specific attributes as the person depicted does not exist in reality and definite statements cannot be made. {\color{black} This is especially important as studies have shown that there are gaps between self-reported and genomic ancestry ethnicity \cite{Mersha2015}, self-reported and observed ethnicity \cite{Saunders2013-gj} and biases and errors in human age estimation \cite{Voelkle2012-fv, doi:10.1098/rsos.180841, DBLP:conf/icb/HanOJ13}}.

In this work, we investigate the demographic and non-demographic diversity of existing synthetic face datasets and also the bias in existing synthetic-based FR models. To do this, we investigate the distribution of head pose, gender, ethnicity, and age in three recent synthetic datasets as well as on the authentic datasets used to train the synthetic data generators using attribute predictors. We also investigate the bias of the synthetic-based FR models regarding gender, age, ethnicity, and pose in comparison to an authentic data-based baseline model. Our results show, that the synthetic face generators generate data similarly, in terms of diversity, to their training data and that the models trained on synthetic models suffer higher bias than the model trained on authentic data, motivating new solutions for {\color{black} more diverse and better synthetic data generators or bias mitigation techniques in synthetic FR models.}

\vspace{-1mm}
\section{Related Work} %Naser Done
\vspace{-2mm}
\subsection{Synthetic Data in Face Biometrics}
\vspace{-1mm}
In recent years, utilizing synthetic data for face biometric tasks has become quite popular \cite{DBLP:conf/icb/BoutrosHSRD22, DBLP:conf/fgr/BoutrosKFKD23, DBLP:conf/iccv/QiuYG00T21, Kolf_2023_CVPR} as ethical and legal requirements have to be met for biometric data using, sharing, and collection {\color{black} in several jurisdictions (such as the European Union \cite{eu-2016/679}). Although the legal requirements may vary culturally and geographically, the recent ethics guidelines of established international venues, such as the International Conference on Computer Vision (ICCV)\footnote{https://iccv2023.thecvf.com/ethics.guidelines-362600-2-25-26.php} or the Conference on Computer Vision and Pattern Recognition (CVPR)\footnote{https://cvpr.thecvf.com/Conferences/2024/EthicsGuidelines}, also require special considerations when publishing or using databases containing personal data}. Moreover, the collection and annotation of large-scale authentic data are expensive and time-consuming and might lead to datasets with low diversity. {\color{black} On the issue of identity relation between synthetic data and the authentic data used to train the generators, Boutros et al. \cite{DBLP:conf/icb/BoutrosHSRD22} have shown close to no relation.}

Recently, a number of works \cite{DBLP:conf/icb/BoutrosHSRD22, DBLP:conf/fgr/BoutrosKFKD23, DBLP:conf/iccv/QiuYG00T21, Kolf_2023_CVPR} proposed the use of privacy-friendly synthetic data to train FR models as an alternative to privacy-sensitive authentic data. Qiu et al. \cite{DBLP:conf/iccv/QiuYG00T21} proposed a synthetic-based FR model, namely SynFace, that utilized synthetic face images generated by attribute-conditional GAN, DiscoFaceGAN \cite{DBLP:conf/cvpr/DengYCWT20}, to train FR models. Each of the synthetic identities in the proposed approach is generated by fixing the identity condition and randomizing the attribute conditions i.e. pose, illumination, and expression.
SynFace \cite{DBLP:conf/iccv/QiuYG00T21} also analyzed the performance gap between synthetic and authentic images as training data and identified poor intra-class variations and the domain gap as possible reasons for verification performance differences. To mitigate this, SynFace proposed to use identity and domain mixup, where identity mixup refers to interpolating between two identities and domain mixup refers to interpolating between authentic and synthetic data in the training data. USynthFace \cite{DBLP:conf/fgr/BoutrosKFKD23} also utilized attribute-conditional GAN to train an FR model in an unsupervised manner. UsynthFace proposed a contrastive learning framework that is trained to maximize the distance between two augmented synthetic images of the same synthetic instance. 
To achieve that,  USynthFace proposed a large set of geometric and color transformations as well as a GAN-based augmentation for their contrastive learning framework. SFace \cite{DBLP:conf/icb/BoutrosHSRD22} and IDNet \cite{Kolf_2023_CVPR} proposed synthetic-based FR models based on class conditional GANs. Each of the synthetic identities in SFace and IDNet is generated by fixing the class label and randomizing the latent code. 
SFace proposed to train StyleGAN-ADA under class conditional settings on CASIA-WebFace. SFace also proposed to improve the synthetic-based FR performances by transferring the knowledge from a model trained on authentic data to the model trained on synthetic data without compromising the authentic identities. Unlike SynFace and UsynthFace, the intra-class variations in SFace are not limited to a predefined set of attributes. However, the generated data by SFace suffers from low identity distinctiveness. IDNet very recently extended SFace by incorporating identity information in the GAN training, aiming at enhancing identity discrimination in the generated data. 
DigiFace-1M \cite{DBLP:conf/wacv/BaeGBHCVCS23} proposed synthetic-based FR, where the synthetic images are generated by rendering digital faces using a computer graphics pipeline. Each synthetic identity in DigiFace is created by randomizing the facial geometry, texture, and hairstyle. Although DigiFace-1M achieved relatively competitive FR verification performances, the generation process is computationally expensive, and the generated images do not match the quality and realistic appearance of authentic images.
Most recently, IDiff-Face \cite{Boutros_2023_ICCV} and ExFaceGAN \cite{ExFaceGAN} were proposed, leading to more realistic face variations and huge advancement in the performance of the synthetic-based FR.
%In the area of FR, Qiu et al. \cite{DBLP:conf/iccv/QiuYG00T21} analyzed the performance gap between synthetic and authentic images as training data and identified poor intra-class variations and the domain gap as possible reasons for verification performance differences. To mitigate this, they proposed to use identity and domain-mixup, where identity mixup refers to interpolating between two identities and domain mixup refers to interpolating between authentic and synthetic data in the training data. Boutros et al. \cite{DBLP:conf/icb/BoutrosHSRD22} proposed to use knowledge distillation to guide a student model trained on synthetic data using a teacher model trained on authentic data to close the performance gap. In 2022, Boutros et al. \cite{DBLP:conf/fgr/BoutrosKFKD23} introduced the concept of unsupervised synthetic FR by generating only a single image of a synthetic identity and use extensive augmentation techniques to create the intra-class variations. In contrast, Kolf et al. \cite{Kolf_2023_CVPR} proposed IDNet, a three player generative adversarial network (GAN), that enables the incorporation of identity information in the face synthesis. 

Besides FR, synthetic face data has also been used for other biometric-related tasks, such as 3D face reconstruction \cite{DBLP:conf/3dim/RichardsonSK16}, presentation attack detection \cite{Fang_2023_CVPR}, morphing attack detection \cite{Damer_2022_CVPR}, FR model quantization \cite{9955645} or face image manipulation \cite{Plesh_2023_CVPR}. In contrast to Fu et al. \cite{10157014}, who investigated the diversity in terms of face image quality of synthetic face images and how they relate to the training images, we investigate the diversity and bias regarding specific demographic and non-demographic attributes.

\vspace{-1mm}
\subsection{Bias in Face Recognition}
\vspace{-1mm}
Exact definitions of bias, implications, and its causes vary between sources. A common understanding is that it relates to differences in performance ratings that are influenced by a particular sub-population \cite{9150783}. Several studies showed that the recognition performance of females is weaker than the performance of male faces when using FR trained on authentic data \cite{DBLP:conf/bmvc/AlbieroB20, DBLP:conf/icb/AlbieroZB20}. Regarding age, studies analyzing the impact of age demonstrated a lower biometric performance for children, than for adults \cite{DBLP:conf/icb/DebN018, DBLP:conf/cvpr/SrinivasRMBK19}. Research investigating the impact of ethnicity showed that faces of under-presented ethnicities perform worse \cite{DBLP:journals/pami/HuangLLT20}. In a comprehensive study, Terhörst et al. \cite{9534882} expanded bias also to non-demographics attributes, such as expression, pose, or illumination. 

To the best of our knowledge, no work so far investigated the bias in synthetic FR models and the diversity of the generated data regarding demographic and non-demographic attributes.

Besides FR, bias can also be observed in other biometric tasks, such as presentation attack detection \cite{DBLP:journals/corr/abs-2209-09035, DBLP:conf/eusipco/FangDKK20}, face image quality assessment \cite{DBLP:conf/icb/TerhorstKDKK20}, biometric systems explanations \cite{DBLP:journals/corr/abs-2304-13419}, or face detection \cite{DBLP:conf/fgr/MittalTMVS23}. 

\vspace{-2mm}
\section{Investigation Methodology \& Setup} %Naser Done
\vspace{-2mm}
In this section, we describe our investigation methodology {\color{black} and the investigation setup}. We start with the approach we use to analyze the diversity of authentic FR datasets and synthetic FR datasets. {\color{black} We describe the utilized datasets, synthetic face generators, and attribute predictors. After that, we describe our approach to investigate the bias of FR models trained on synthetic faces, including the FR models and evaluation methods used.}

\subsection{Diversity Investigation}
To investigate the diversity of the datasets, we use different high-performing attribute predictors to predict the three demographics attributes gender, age, and ethnicity, and the non-demographic attributes head pose. We then report the different distributions of the attributes of the different authentic and synthetic datasets. 

The investigation of the diversity of the authentic datasets is also of special interest, as the investigated datasets are used to train the generative models that then create the synthetic datasets. Investigating this allows possible insights into how the distribution changes based on the utilized training data, as some might assume that the generative model in general follows the distribution that has been used to train it.

%In addition, we also investigate the consistency and diversity of the investigated attribute (gender, ethnicity, age, head pose, {\color{blue} and face expression}) within identities using a novel metric named Intra-Identity Attribute Consistency Ratio (IIACR). This metric captures how consistent a specific attribute is for the identities of a dataset. This is of specific interest as in contrast to authentic data, the identity constraint that can be observed in synthetic data is less strict than authentic data, which may cause the observable domain gap \cite{DBLP:conf/iccv/QiuYG00T21}. In synthetic data, as the individual is not real, there is no ground truth about the gender, age, or ethnicity as this might vary, also based on subjective interpretation. These features may be less prominent and may contradict each other in a single image. Regarding attribute diversity and consistency within an identity, we expect attributes such as gender or ethnicity having a high consistency as only in a few cases, do individuals change these attributes regarding their appearance in authentic data. For the attributes of head pose, {\color{blue} face expression} and age, we expect less consistency and assume that in terms of intra-class variance, a higher diversity is beneficial.

\vspace{-2mm}
\subsubsection{Attribute Predictors}
\vspace{-1mm}
Since the investigated datasets do not provide human-labeled attributes, labeling large datasets is expensive, and there is also no factual self-reported ground truth in synthetic images regarding specific attributes, we utilize attribute predictors to get automatic attribute labels. In our experiments, we limit ourselves to the most investigated attributes gender, age, ethnicity, and head-pose \cite{DBLP:journals/corr/abs-2003-02488}. {\color{black} The results based on four additional well-established \cite{DBLP:conf/fgr/BernalC23, DBLP:conf/uist/EvirgenC22, DBLP:journals/pvldb/RomeroHPKZK22, 10219710} attribute estimator (including another non-demographic attribute face emotion) based on an open source project \cite{9659697, serengil2020lightface} are provided in the supplementary material due to space.} For each of the attributes, we utilize different well-performing attribute predictors.

The \textbf{gender predictor} \cite{DBLP:journals/ijcv/RotheTG18} is based on VGG-16 architecture \cite{DBLP:journals/corr/SimonyanZ14a} and was pre-trained on ImageNet \cite{DBLP:journals/ijcv/RussakovskyDSKS15} and fine-tuned on the IMDb-Wiki dataset \cite{DBLP:journals/ijcv/RotheTG18}. It achieved a classification accuracy of $88.50\%$ on the Balanced-Faces-in-the-Wild (BFW) \cite{nmsj-df12-22} dataset. We decided to limit our prediction classes to two genders ($male$, $female$), being aware that there are more than two genders people identify themselves with. 

The \textbf{age predictor} we utilized a support vector machine (SVM) trained on feature embeddings extracted using ElasticFace-Arc \cite{DBLP:conf/cvpr/BoutrosDKK22} from the Adience \cite{DBLP:conf/iccvw/SamekBLM17} dataset. It achieved a mean accuracy of $60.51\% \pm 2.28$ in a five cross-fold evaluation setup on Adience \cite{DBLP:conf/iccvw/SamekBLM17}, which is comparable to other works on the hard Adience dataset \cite{DBLP:conf/btas/TerhorstHKZDKK19, DBLP:conf/fusion/BoutrosDTKK19, DBLP:conf/fusion/TerhorstHKDKK19, DBLP:journals/ijcv/RotheTG18}. The Adience dataset provides 8 classes, which are defined as: $(0,2)$, $(4,6)$, $(8,12)$, $(15,20)$, $(25,32)$, $(38,43)$, $(48,53)$, and $(60,100)$. 

As the \textbf{ethnicity predictor}, we also utilize an SVM trained on feature embeddings extracted using ElasticFace-Arc \cite{DBLP:conf/cvpr/BoutrosDKK22}. The images to create the feature embeddings are taken from BUPT-Balancedface \cite{DBLP:journals/pami/WangZD22} dataset, which provides a  large set of nearly equally distributed ethnicities. The datasets distinguished between the ethnicities $African/Black$, $Asian$, $Caucasian/White$, and $Indian$. The SVM is trained on randomly selected 10\% of the data while keeping the distribution of the ethnicities equal. To evaluate the performance, we test on 1,300 images also randomly selected from BUPT-Balanceface while ensuring that the identities from the training set are not part of the test set. On this test set, we achieved an accuracy of $90.91\%$.

As the \textbf{head-pose predictor} we use Hopenet \cite{DBLP:conf/cvpr/RuizCR18}, which is one of the publicly available top-performing head pose estimators. It uses a multi-loss convolutional neural network and predicts intrinsic Euler angles (yaw, pitch, and roll) of the head pose. For simplification, we only consider yaw and evaluate on the Annotated Facial Landmark in the Wild (AFLW) \cite{DBLP:conf/iccvw/KostingerWRB11} dataset, which results in a mean absolute error of the predictor of 8.26. This implies that the predicted yaw angle differs from the real angle by about 8.26. Since we mainly care about the general head pose, we divide the obtained yaw into five classes: $0\degree$ (frontal), $22.5\degree$, $45\degree$, $67.5\degree$, and $90\degree$ (profile) based on the yaw angle. 

\vspace{-2mm}
\subsubsection{{\color{black} Diversity Evaluation Metric}}
\vspace{-1mm}
To measure and evaluate the diversity of the authentic and synthetic training data, we report the overall data distribution in terms of gender, age, ethnicity, and head pose as predicted by our attribute predictors. Since in some cases, different samples of the same identity might have conflicting attribute predictions, we also want to gain insights into the intra-identity distribution of the attributes. Differences in these attributes might be natural (e.g. aging process) or due to less distinct features. We propose the novel \textit{Intra-Identity Attribute Consistency Ratio (IIACR)}. The IIACR is calculated as:
\begin{equation}
\small
    IIACR = \frac{1}{N}\sum^N_{n=1}\frac{max_{\alpha \in S}(m_{n,a})}{m_n},
\end{equation}
where $N$ is the number of identities, $S$ is the set of different classes for an attribute, $\alpha$ is a class of $S$, $m_n$ is the total amount of images of an identity $n$, and $m_{n,a}$ is the amount of images of an identity $n$ that belongs to class $\alpha$. The IIACR, therefore, provides insights into the consistency of an attribute within the images of an identity. For some attributes such as gender or ethnicity, this value is expected to be 1, as individuals rarely change their gender or ethnicity. In synthetically generated face images, where the creators rather aim at preserving or creating the synthetic identity than maintaining the similar attribute for the different samples of a synthetic identity, this might vary more often. For other attributes, such as head pose and age, a lower value might be favorable, as this indicates more variety in the face data. In our experiments, we randomly select 1\% of the identities and calculate the IIACR based on all available images of the randomly selected individuals. On USynthFace-400k, we only applied it to the GAN-augmented samples, as the geometric and color transformation had been done online.

\vspace{-1mm}
\subsection{Bias Investigation}
\vspace{-1mm}
To investigate the bias in FR models trained on synthetic or authentic data, we analyze performance differences depending on the investigated demographic attributes (gender, cross-age, ethnicity) and non-demographic attributes (head pose). To obtain these performance differences, we evaluate different datasets that make a distinction regarding the specific attributes, e.g. we compare between the verification accuracy of a female subset to that of a male subset. On all benchmarks, we follow the provided evaluation protocol to produce reproducible results. 

Finally, we combine the results from the diversity and consistency analysis with the results from the bias analysis to investigate the influence of the synthetic or authentic training data on the different verification performances.

\vspace{-3mm}
\subsubsection{{\color{black} Bias Evaluation Datasets \& Metric}}
\vspace{-2mm}
For the evaluation in terms of bias, we utilize six different datasets. The different datasets provide different labels to create attribute-based subsets and are often used in bias studies \cite{9150783, DBLP:conf/iccv/WangDHTH19}. For the evaluation in terms of gender bias, we report the performance on Balanced-Faces-in-the-Wild (BFW) \cite{nmsj-df12-22}. To investigate the ethnicity bias, we evaluate the different models on BFW \cite{nmsj-df12-22} as well as Racial-Faces-in-the-Wild (RFW) \cite{DBLP:conf/iccv/WangDHTH19}. To investigate the performance difference on images with different ages, we compare LFW \cite{LFWTech} (smaller age gap) with Cross-Age LFW \cite{DBLP:journals/corr/abs-1708-08197} (larger age gap) as they are based on the same data but instead of random comparisons (LFW), the Cross-Age LFW dataset consist of genuine and imposter pairs with higher age gaps. For the head-pose performance difference, we compare the performance on CFP-FF (frontal-frontal) \cite{DBLP:conf/wacv/SenguptaCCPCJ16} to CFP-FP (frontal-profile) \cite{DBLP:conf/wacv/SenguptaCCPCJ16} which are also based on the same data but provide a different pose evaluation scenario, were the pairs of CFP-FF both show frontal images, while in the CFP-FP datasets, one face image is a profile image. 

To investigate the bias in synthetic FR models, we report the verification accuracy on the different subsets following the defined protocol of each dataset, as well as the mean accuracy (mAcc) and the standard deviation (STD), similar to other bias analyses works \cite{DBLP:journals/corr/abs-1911-10692, DBLP:journals/corr/abs-2304-02284, DBLP:conf/cvpr/WangD20}. Furthermore, we also report the Skewed Error Ratio (SER) \cite{DBLP:journals/corr/abs-1911-10692}. Error skewness is computed by the ratio of the highest error rate to the lowest error rate among different attributes and is therefore calculated as:
\begin{equation}
\small
    SER = \frac{max_{a}Err(a)}{min_{b}Err(b)}
\end{equation}
where $a$, $b$ are classes of the investigated attribute.

\vspace{-1mm}
\subsection{Face Recognition Models \& Datasets} % Marco Done
\label{frmodels}
\vspace{-1mm}
To investigate the bias in FR models trained on synthetic data, we utilized three different recently proposed models trained on their associated synthetic training data. The utilized models are SFace$_{synth}$ \cite{DBLP:conf/icb/BoutrosHSRD22}, SynFace \cite{DBLP:conf/iccv/QiuYG00T21}, and USynthFace \cite{DBLP:conf/fgr/BoutrosKFKD23}. We chose these models because they provide state-of-the-art synthetic FR and are of different natures in their approach. All models used are publicly available and have been released by the authors of the respective works.
% how to achieve FR based on synthetic data. 

As a \textbf{Baseline}, we investigate the bias in FR models trained on authentic data, ResNet50 \cite{DBLP:conf/cvpr/HeZRS16} trained on CASIA-WebFace \cite{DBLP:journals/corr/YiLLL14a} with CosFace loss \cite{DBLP:conf/cvpr/WangWZJGZL018}. This model is considered in our evaluation as it was used by the utilized synthetic-based FR models \cite{DBLP:conf/icb/BoutrosHSRD22, DBLP:conf/iccv/QiuYG00T21} as a baseline for comparing their verification performances with the model trained on authentic data.
%To compare with a FR model trained on authentic data, we make use of a baseline model trained on CASIA-WebFace \cite{DBLP:journals/corr/YiLLL14a} as it was considered as the authentic-based baseline in the considered synthetic-based FR works \cite{DBLP:conf/icb/BoutrosHSRD22, DBLP:conf/iccv/QiuYG00T21, DBLP:conf/fgr/BoutrosKFKD23}. 

\textbf{SFace$_{synth}$} \cite{DBLP:conf/icb/BoutrosHSRD22} is a model trained on the SFace-60 dataset. 
The model architecture is ResNet-50 \cite{DBLP:conf/cvpr/HeZRS16} trained with CosFace loss \cite{DBLP:conf/cvpr/WangWZJGZL018}. During the training phase, SFace proposed to transfer the knowledge from a model trained on authentic data, CASIA-WebFace \cite{DBLP:journals/corr/YiLLL14a}. This aims at guiding the synthetic FR model to learn to produce feature representations that are similar to the ones learned by the model trained on authentic data. It should be noted that the training process of SFace does not include any authentic data.
The SFace training dataset consists of 634,320 images of 10,572 identities (60 images per identity). The images have been generated by a StyleGAN2-ADA \cite{DBLP:conf/nips/KarrasAHLLA20} conditionally trained on CASIA-WebFace \cite{DBLP:journals/corr/YiLLL14a}. 

The authentic \textbf{CASIA-WebFace} \cite{DBLP:journals/corr/YiLLL14a} dataset used to train the FR baseline and the SFace generative model consists of 494,414 images of 10,575 different identities. The images in CASIA-WebFace have been collected semi-automatically from the web. The authors of the datasets did not make any statement on the diversity of their dataset, but it is regularly used to train FR systems \cite{DBLP:conf/cvpr/BoutrosDKK22, DBLP:conf/cvpr/DengGXZ19, DBLP:conf/cvpr/Kim0L22}, especially when compared to synthetic-based FR \cite{DBLP:conf/icb/BoutrosHSRD22, DBLP:conf/iccv/QiuYG00T21}.

\textbf{SynFace} \cite{DBLP:conf/iccv/QiuYG00T21} was trained by the authors on the Syn\_10k\_50 dataset using identity-mixup. The backbone architecture is ResNet-50 \cite{DBLP:conf/cvpr/HeZRS16} trained with  ArcFace loss \cite{DBLP:conf/cvpr/DengGXZ19}. \textbf{Syn\_10k\_50} \cite{DBLP:conf/iccv/QiuYG00T21} consists of 500,000 synthetic images of 10,000 different identities.  Syn\_10k\_50 utilized DiscoFaceGAN \cite{DBLP:conf/cvpr/DengYCWT20} to generate the synthetic face images. DiscoFaceGAN is an attribute conditional GAN model trained on the FFHQ dataset \cite{DBLP:conf/cvpr/KarrasLA19}. Synthetic identities in Syn\_10k\_50 are generated by fixing the identity condition and randomly sampling latent variables from the standard
normal distribution for expression, pose, and illumination. The FFHQ dataset contains 70k images collected from Flickr and encompasses variation in ethnicity, age, image background, and accessories \cite{DBLP:conf/cvpr/KarrasLA19}.

\textbf{USynthFace} \cite{DBLP:conf/fgr/BoutrosKFKD23} is an unsupervised FR model trained with unlabeled synthetic data.  USynthFace FR model architecture is ResNet-50 \cite{DBLP:conf/cvpr/HeZRS16} trained with contrastive learning. Similar to the SynFace model, USynthFace utilized a DiscoFaceGAN trained on FFHQ to generate the synthetic dataset, USynthFace-400k. The USynthFace-400k dataset consists of 400,000 images of 400,000 synthetic identities.

In the diversity analysis we analyze two authentic datasets CASIA-Webface \cite{DBLP:journals/corr/YiLLL14a}, FFHQ \cite{DBLP:conf/cvpr/KarrasLA19} that had been used during the generator training, and the three different synthetic datasets, SFace-60 \cite{DBLP:conf/icb/BoutrosHSRD22}, Syn\_10k\_50 \cite{DBLP:conf/iccv/QiuYG00T21} and USynthFace-400k \cite{DBLP:conf/fgr/BoutrosKFKD23} that has been created by the generators. In the bias analysis, we investigate the bias of the authentic baseline model and the three synthetic models, SFace$_{synth}$, SynFace, and USynthFace.

\vspace{-1mm}
\section{Results} %Naser Done % Marco Done
\vspace{-2mm}
In this section, we present the results of our investigation. First, we provide the results of our data diversity investigation starting with the distributions of gender, ethnicity, age, and head pose on the five different datasets. Later on, we present and discuss the results of the intra-identity attribute consistency analysis, to evaluate how consistent or diverse the different attributes are in an authentic dataset and synthetic dataset. {\color{black} The diversity distributions for the predictions of four additional attribute estimators are provided in the supplementary material.}

\begin{figure} [t]
    \centering
    \includegraphics[width=0.85\columnwidth]{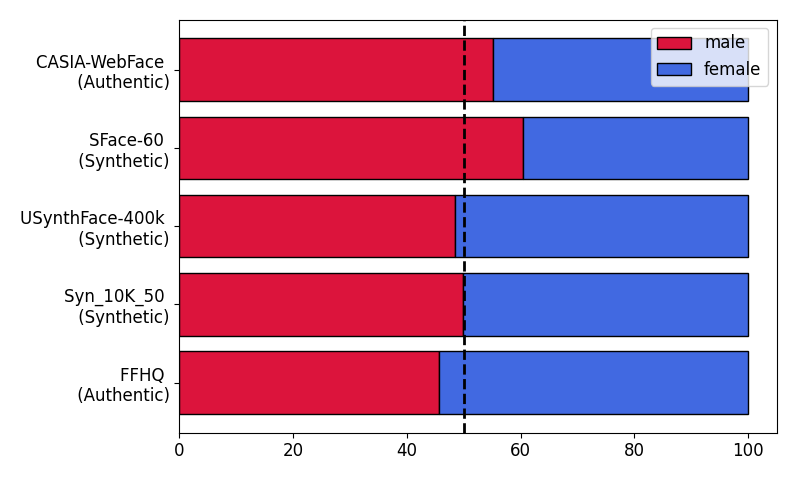}
    \caption{\textbf{Gender Distribution:} The dotted line indicates an equal distribution. While the FFHQ dataset, USynthFace-400k dataset, and Syn\_10K\_50 dataset are nearly balanced, the imbalance of the gender distribution in the SFace-60 dataset increased in contrast to its generator training data, CASIA-WebFace. This might indicate a bias regarding generating individuals from the majority class.}
    \label{fig:genderdist}
    \vspace{-4mm}
\end{figure}

\begin{figure}[t]
    \centering
    \includegraphics[width=0.85\columnwidth]{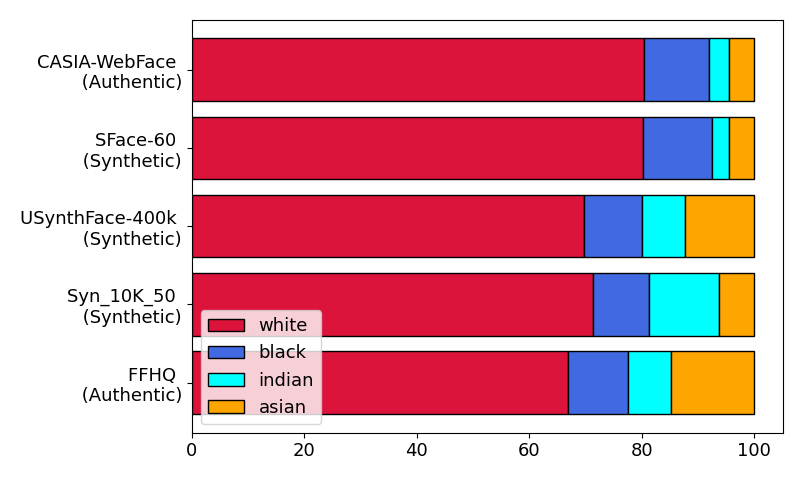}
    \caption{\textbf{Ethnicity Distribution:} All datasets show a high ethnicity imbalance. The synthetic datasets seem to inherit the general distribution regarding ethnicities from the authentic training set to train the generators. (FFHQ was training data for the Syn\_10K\_50 generator and the USynthFace-400k generator, CASIA-WebFace was used to train and generate the SFace-60 dataset.}
    \label{fig:ethndist}
    \vspace{-6mm}
\end{figure}

\vspace{-1mm}
\subsection{Data Diversity}
\vspace{-1mm}
\subsubsection{Attribute Distribution}
\vspace{-1mm}
\begin{table*}[]
\centering
%\small
\resizebox{0.7\linewidth}{!}{%
\begin{tabular}{l|cc|cccc} \hline \hline
\multicolumn{1}{c|}{Dataset}  & \multicolumn{2}{c|}{Gender} & \multicolumn{4}{c}{Ethnicity}                    \\ \hline 
                                                                    & Male        & Female       & African/Black & Asian & White/Caucasian & Indian \\ \hline
\begin{tabular}[c]{@{}l@{}}CASIA-WebFace (auth.)\end{tabular} & 55.23       & 44.77        & 11.49         & 4.45  & 80.45           & 3.61   \\
\begin{tabular}[c]{@{}l@{}}SFace-60 (syn.)\end{tabular}      & 66.52       & 39.48        & 12.20         & 4.44  & 80.20           & 3.16   \\
\begin{tabular}[c]{@{}l@{}}USynthFace (syn.)\end{tabular}    & 48.56       & 51.44        & 10.44         & 12.26 & 69.64           & 7.66   \\
\begin{tabular}[c]{@{}l@{}}Syn\_10K\_50 (syn.)\end{tabular}  & 49.80       & 50.20        & 9.93          & 6.21  & 71.40           & 12.46  \\
\begin{tabular}[c]{@{}l@{}}FFHQ (auth.)\end{tabular}          & 45.62       & 54.38        & 10.70         & 14.85 & 66.87           & 14.85  \\ \hline \hline 
\end{tabular}}
\caption{\textbf{Distribution of Gender and Ethnicity in \%:} Similar to Figure \ref{fig:genderdist}, the values show that the synthetic SFace-60 is more imbalanced than its authentic origin dataset CASIA-WebFace. Regarding the ethnicity distribution, the synthetic datasets inherit the general balance from their authentic origin dataset (see also Figure \ref{fig:ethndist}).}
\label{tabl:gendereth}
\vspace{-3mm}
\end{table*}

\begin{table*}[]
\centering
\resizebox{\linewidth}{!}{%
\begin{tabular}{l|cccccccc|ccccc} \hline \hline
Dataset                                                             & \multicolumn{8}{c|}{Age}                                                           & \multicolumn{5}{c}{Head Pose} \\ \hline 
                                                                    & (0,2) & (4,6) & (8,12) & (15,20) & (25,32) & (38,43) & (48,53) & (60,100) & 0\degree  & 22.5\degree  & 45\degree  & 67.5\degree  & 90\degree \\ \hline 
CASIA-WF. (auth.) &  0.01      & 0.31       & 1.79        & 5.21         & 67.74         & 19.02         & 2.97         & 2.96          & 51.01   & 36.12      & 8.14    & 3.65      & 1.06   \\
SFace-60 (syn.)      & 0.00       & 0.18       & 1.10        & 5.16         & 63.30         & 23.92         & 2.99         & 3.36          & 48.37   & 37.07      & 10.21    & 3.88      & 0.47   \\
USynthFace (syn.)   & 1.87       & 6.70       & 3.79        & 14.55         & 53.34         & 18.32         & 1.17         & 0.57          & 59.85   & 35.78      & 4.18    & 0.21      & 0.00   \\
Syn\_10K\_50 (syn.)  & 0.74       & 6.27       & 4.43        & 15.73         & 54.62         & 17.34         & 0.69         & 0.19          & 63.97   & 33.71      & 2.26    & 0.06      & 0.00   \\
FFHQ (auth.)          & 2.88       & 7.40       & 5.45        & 8.70         & 49.80         & 17.10         & 4.65         & 4.03          & 59.89   & 36.03      & 3.79    & 0.28      & 0.02  \\ \hline \hline 
\end{tabular}}
\caption{\textbf{Distribution of Head Pose and Age in \%:} Similar to Figure \ref{fig:agedist}, the percentages in the Table show that there is a high imbalance towards the age ranges (25,32) and the adjacent age ranges in all datasets. The infant and elderly classes are under-represented in all datasets, but the imbalance increased when using the generators trained on FFHQ to create the USynthFace and the Syn\_10K\_50 datasets. Regarding head pose, the CASIA-WebFace and also SFace-60 provide higher diversity, but the diversity is reduced in the synthetic training dataset as fewer profile or next-to-profile images are present in the dataset (see also Figure \ref{fig:hpdist}).}
\label{agehp}
\vspace{-5mm}
\end{table*}

The distribution of the gender attribute is visualized in Figure \ref{fig:genderdist} and the percentages are also shown in Table \ref{tabl:gendereth}. The distribution of male and female individuals in the dataset is close to even, while the authentic FFHQ dataset shows more female individuals, the other authentic dataset, CASIA-WebFace shows more male individuals. SFace-60, the synthetic dataset which has been created utilizing a generative model trained on CASIA-Webface revealed a higher imbalance regarding gender, which might indicate that the generative model tends to create samples from its majority class regarding gender distribution. This is especially interesting, as the synthetic identities of the SFace-60 dataset have been created based on the original CASIA-WebFace identities. The generative model, therefore, at least in some cases, flipped the gender of some images or identities in their synthetic counterparts from female to male.

The ethnicity distribution is shown in Figure \ref{fig:ethndist} and the numerical values in Table \ref{tabl:gendereth}. The figure shows that there is a highly imbalanced distribution in the authentic, but also in the synthetic face datasets. White/Caucasian individuals are highly over-represented while other ethnicities are under-represented. One can also note that the synthetic datasets follow, to some degree, the overall distribution of the respective generator training data of the generator. The generative model trained on the slightly less imbalanced FFHQ dataset led to less imbalanced synthetic datasets USynFace-400k and Syn\_10K\_50 in contrast to the more imbalanced CASIA-WebFace and SFace-60 dataset.

\begin{figure}
    \centering
    \includegraphics[width=0.9\columnwidth]{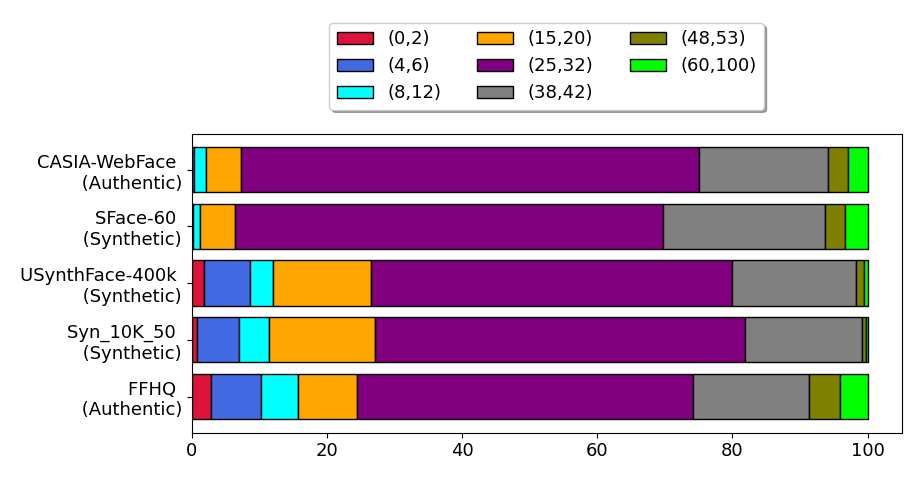}
    \caption{\textbf{Age Distribution:} A relatively higher representation of the age range of (25,32)  and the adjacent age ranges can be observed. The age distribution seems also to be inherited from the authentic datasets, as SFace-60 mimics the distribution of CASIA-WebFace. USynthFace-400k and Syn\_10K\_50 are also roughly similar to FFHQ with some differences in the elderly and infant class, which are more under-represented in the synthetic datasets.}
    \label{fig:agedist}
    \vspace{-5mm}
\end{figure}

The age distribution is presented in Figure \ref{fig:agedist} and the distribution percentages in Table \ref{agehp}. Again, a high imbalance can be seen across the datasets. The majority of the images are classified as showing individuals in the age range of (25-32). Since the age predictor is not perfect, but also generally provides a high one-off accuracy \cite{DBLP:conf/btas/TerhorstHKZDKK19, DBLP:conf/fusion/BoutrosDTKK19}, the high amount of samples in the adjacent age ranges of (15,20) and (38,42) might be explained. The small number of children and adolescents in the data sets is noticeable, although it is also noticeable that, similar to ethnicity, the general distribution of the attributes remains the same in the respective generator training dataset.

\begin{figure}
    \centering
    \includegraphics[width=0.85\columnwidth]{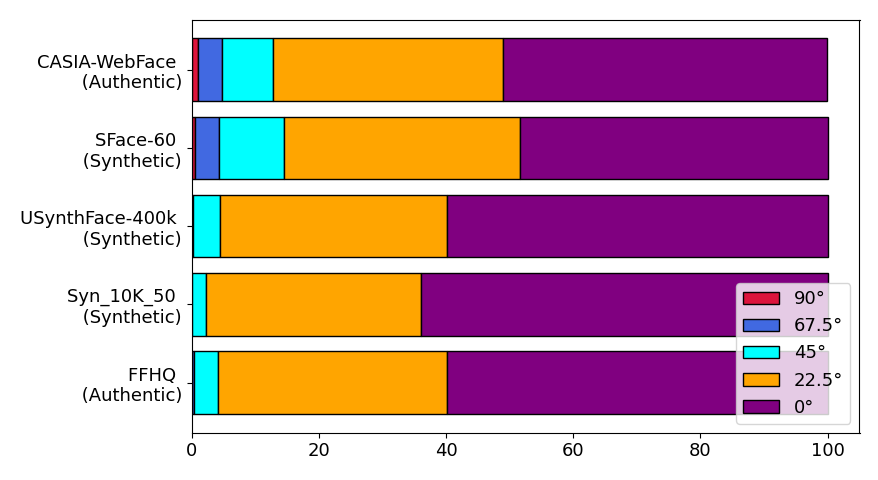}
    \caption{\textbf{Headpose Distribution:} The large majority of the face images in the authentic and synthetic datasets are frontal or near frontal. Profile and near profile images are under-represented and the percentage decreases at least when using the CASIA-WebFace-based generator to create SFace-60 (see also Table \ref{agehp}).}
    \label{fig:hpdist}
    \vspace{-2mm}
\end{figure}

The distribution of the non-demographic head pose attribute is shown in Figure \ref{fig:hpdist} and Table \ref{agehp}. To simplify the evaluation, we neglect the direction of the yaw angle, i.e. we do not differentiate between the right or left profile of the face. From Figure \ref{fig:hpdist} it can be observed, that there is a high imbalance regarding frontal or nearly frontal images (yaw angle of around 22.5\degree) on all datasets. Interestingly, SFace-60 seems to contain more images with a semi-profile view (45\degree) than the original CASIA-WebFace increasing the diversity of the training data. The datasets based on FFHQ show a very similar head pose distribution to the authentic dataset used to train their generator. 

\vspace{-4mm}
\subsubsection{Intra-Identity Attribute Consistency}
\vspace{-2mm}
We also investigate the consistency of our investigated attributes to gain insights into how variable the specific attributes are and how good the synthetic models can maintain these, as they normally aim at preserving synthetic identity information and not the specific attributes. With the IIACR we measure the mean intra-identity attribute consistency over the analyzed subsets of the datasets. For attributes such as gender and ethnicity, we expect it to be beneficial for the FR training, if the IIACR is high, as this means that less noise or domain difference is introduced into the model. For attributes such as age and pose, we assume a lower consistency to be beneficial as this means a higher natural intra-class variability is present in the training dataset.

\begin{table}[]
\centering
\footnotesize
\begin{tabular}{c|cccc} \hline \hline
Dataset         & Gender    & Age       & Ethnicity & Pose      \\ \hline 
CASIA-WF   & 0.95$\pm$0.07 & 0.78$\pm$0.16 & 0.92$\pm$0.11 & 0.57$\pm$0.14 \\ \hline 
SFace-60        & 0.93$\pm$0.10 & 0.67$\pm$0.15 & 0.91$\pm$0.07 & 0.49$\pm$0.10 \\
Syn\_10K\_50    & 0.71$\pm$0.11 & 0.57$\pm$0.13 & 0.73$\pm$0.16 & 0.64$\pm$0.07 \\
USynthFace & 0.92$\pm$0.13 & 0.76$\pm$0.18 & 0.82$\pm$0.18 & 0.60$\pm$0.07 \\ \hline  \hline 
\end{tabular}
\caption{\textbf{Intra-Identity Attribute Consistency Ratio:} SFace-60 and USynthFace-400k show a similar IIACR as CASIA-WebFace regarding gender and ethnicity, indicating that most images of one synthetic individual share the same attribute features. Syn\_10K\_50 shows a different behavior with less consistency, which might be due to the identity-mixup without checking for matching gender or ethnicity. The age and pose IIACR is lower for SFace-60 than for CASIA-WebFace which might indicate a higher intra-class variation of the synthetic data in contrast to the authentic data.}
\label{IIACR}
\vspace{-3mm}
\end{table} 

\begin{table}[]
\centering
\footnotesize
\begin{tabular}{c|cc|rr} \hline \hline
Model   & Male  & Female & mAcc+STD & SER \\ \hline 
Baseline & 93.45 & 91.52  & 92.49$\pm$0.97 & 1.29    \\ \hline 
SFace$_{syth}$ & 92.38 & 89.67  & 91.03$\pm$1.35 & 1.36    \\
SynFace & 80.52 & 77.96  & 79.24$\pm$1.28 & 1.13    \\
USynthFace  & 81.27 & 78.27  & 79.77$\pm$1.50  & 1.16    \\ \hline \hline
\end{tabular}
\caption{\textbf{Gender Bias on BFW:} All the models, including the authentic model performed better on male faces than female faces. While the STD indicates less bias in the authentic model, the SER indicates higher bias in the authentic data.}
\label{genderbias}
\vspace{-3mm}
\end{table}

\begin{table*}[]
\centering
\footnotesize
\begin{tabular}{c|cccc|rr} \hline \hline
               & \multicolumn{6}{c}{RFW}                                      \\ \hline \hline
Model          & African & Asian & Caucasian & Indian & mAcc + STD     & SER  \\  \hline
Baseline & 85.52   & 84.27 & 92.52     & 88.03  & 87.59$\pm$3.15 & 2.10 \\  \hline
SFace$_{syth}$ & 80.17   & 80.93 & 90.15     & 84.32  & 83.89$\pm$3.94 & 2.01 \\
SynFace        & 59.75   & 67.20 & 70.05     & 66.00  & 65.75$\pm$3.76 & 1.34 \\
USynthFace     & 60.25   & 67.67 & 72.40     & 69.13  & 67.36$\pm$4.45 & 1.44 \\  \hline \hline
               & \multicolumn{6}{c}{BFW}                                      \\ \hline \hline
Model               & Black   & Asian & White     & Indian & mAcc + STD     & SER  \\  \hline
Baseline & 92.20   & 87.43 & 95.56     & 90.71  & 91.48$\pm$2.92 & 2.83 \\  \hline
SFace$_{syth}$ & 90.05   & 85.66 & 94.04     & 88.99  & 89.69$\pm$2.99 & 2.41 \\
SynFace        & 74.92   & 72.79 & 79.47     & 75.90  & 75.77$\pm$2.41 & 1.32 \\
USynthFace     & 75.66   & 74.01 & 80.35     & 77.06  & 76.77$\pm$2.33 & 1.32 \\  \hline  \hline
\end{tabular}
\caption{\textbf{FR performance and ethnicity Bias on BFW and RFW:} The authentic-based baseline model as well as the synthetic-based models show a high positive bias regarding Caucasians/Whites. While the STD and SER on RFW indicate higher bias of the synthetic models toward the ethnicities, the results on BFW might indicate that a lower intra-class ethnicity consistency might be beneficial to reduce bias.}
\label{ethbias}
\vspace{-5mm}
\end{table*}

\begin{table*}[]
\centering
\footnotesize
\begin{tabular}{c|cc|rr} \hline \hline
Model   & LFW & Cross-Age LFW & mAcc + STD & SER \\ \hline
Baseline        & 99.38      & 93.22    & 96.30$\pm$3.80 & 10.94    \\ \hline
SFace$_{syth}$  & 99.02      & 92.08    & 95.55$\pm$3.47 & 8.08    \\
SynFace         & 91.77      & 75.18    & 83.48$\pm$8.29 & 3.02   \\
USynthFace      & 91.83      & 76.88    & 84.35$\pm$7.48 & 2.83   \\ \hline \hline
\end{tabular}
\caption{\textbf{Age Bias on LFW and Cross-Age LFW:} All the models perform worse when confronted with comparison pairs of higher age gaps. The absolute performance drop was even higher on the synthetic-based models than the authentic-based baseline. SER is consistently lower (better) for synthetic-based models.}
\label{agegap}
\vspace{-5mm}
\end{table*}

\begin{table}[]
\centering
\footnotesize
\begin{tabular}{c|cc|rr} \hline \hline
Model   & F-F & F-P & mAcc + STD & SER \\ \hline
Baseline  & 99.33      & 95.84    & 97.59$\pm$1.74 & 6.21    \\ \hline
SFace$_{syth}$  & 98.80      & 91.90    & 95.35$\pm$3.45 & 6.75    \\
SynFace         & 90.33      & 74.53    & 82.43$\pm$7.90 & 2.63    \\
USynthFace      & 90.34      & 78.20    & 84.27$\pm$6.07 & 2.26   \\ \hline \hline
\end{tabular}
\caption{\textbf{Headpose Bias on CFP-FF (frontal-frontal) and CFP-FP (frontal-profile)}: The results show that cross-pose verification accuracy is worse on every model, authentic and synthetic-based. The absolute performance decrease is higher on the synthetic-based models than on the authentic baseline.}
\label{hpgap}
\vspace{-5mm}
\end{table}

Table \ref{IIACR} shows the IIACR values for the authentic CASIA-WebFace dataset as a reference dataset for an authentic dataset and the three synthetic datasets. The high values of 0.95 for gender and 0.92 for ethnicity indicate the high consistency of the gender and ethnicity attributes on authentic data. SFace-60, which utilizes the identities of CASIA-WebFace to create its synthetic identities also achieves a high consistency in these attributes, which indicates a lower domain gap. The IIACR for Syn\_10K\_50 dataset shows a lower IIACR regarding gender and ethnicity which might be due to the identity-mixup during the face synthesis which might mix up individuals of different genders or ethnicities, leading to contradicting face features in different images.

To summarize our investigation of the diversity of the training dataset showed, that the overall distribution of the authentic training dataset is to some degree inherited by the synthetic training data, depending on which data the generator has been trained. In some cases, the imbalance is even increased after the face synthesis. The investigation on the intra-identity consistency showed that most synthetic datasets show a high consistency regarding fixed attributes such as gender and ethnicity, while having a similar variability regarding changing attributes such as age and pose.

\vspace{-1mm}
\subsection{Bias in Synthetic-based FR Models}
\vspace{-1mm}
In this subsection, we investigate the bias in the decisions of synthetic-based FR models in comparison to a baseline model trained on authentic data. We investigate this on gender, ethnicity, age, and head pose. To evaluate this, we report the subset-specific accuracy, the mean accuracy (mAcc) including standard deviation (STD), and the SER, following \cite{DBLP:journals/corr/abs-1911-10692, DBLP:journals/corr/abs-2304-02284, DBLP:conf/cvpr/WangD20}. {\color{black} Following the "Rule of 30" and its extensions \cite{DODDINGTON2000225}, in most cases, a confidence level of 90\% with a percent relative error of 10\% is achieved regarding the reported accuracies and errors. Only the Baseline and the SFace model are below the required number of errors on the LFW and FF dataset to achieve this confidence and percent relative error. Given the error rates in the Tables \ref{genderbias}, \ref{ethbias}, \ref{agegap} and \ref{hpgap} and the level of confidence for the percent relative bias, the bias in most cases is statistically significant.}

Table \ref{genderbias} presents the verification accuracy on the gender subsets of the BFW dataset. In all cases, the performance is worse on female than male faces, which is consistent with previous findings \cite{DBLP:conf/bmvc/AlbieroB20, DBLP:conf/icb/AlbieroZB20}. While the performance of the SFace$_{synth}$ is competing with the Baseline model, the other models perform worse.

To analyze ethnicity bias, we provide the verification accuracy, the mean accuracy, standard deviation, and SER on the ethnicity split on RFW and BFW in Table \ref{ethbias}. In the reported values we can observe an ethnicity bias as the Caucasian/White ethnicity class is the best-performing attribute class regarding verification accuracy across all investigated models. Similar to the observation on the gender bias, we observe that the STD of the synthetic model on the RFW dataset is higher in contrast to the STD of the authentic models, while the SER is lower, indicating a lower bias. Interestingly, on the BFW dataset, the SER and the STD are lower for SynFace and USynthFace, meaning lower bias. A possible explanation might be, that due to the less constrained identity attributes and a higher inconsistency regarding ethnicity within an identity in the synthetic data (see \ref{IIACR}), the synthetic models are less biased.

The results of the age bias evaluation are presented in Table \ref{agegap}. In contrast to the other datasets, the setup on age is slightly different, as the distinction is not made between different age groups, but between comparisons of more similar age groups and cross-age comparisons. The results show, that similar to authentic models, the performance on higher age gaps between compared samples reduced the performance also for synthetic models. The higher variation in the intra-identity age attribute might also be beneficial, as a lower SER can be reported for the synthetic models than the authentic baseline. 

Finally, the results on variation in head pose are presented in Table \ref{hpgap}. We compare two scenarios: Frontal-Frontal and Frontal-Profile comparison pairs. All the models perform worse on the frontal-profile scenario indicating a higher challenge and biased behavior. Analyzing the results of the synthetic models shows that the performance difference between frontal-frontal and frontal-profile increased in contrast to the authentic model.

To summarize, in total we observed similar biases in the synthetic FR models as in the authentic FR model, which motivates the use and development of bias mitigation techniques also on synthetic-based FR models. The intra-identity attribute consistency which might be lower in synthetic data due to fewer constraints might be beneficial to reduce bias.

\vspace{-1mm}
\section{Conclusion}
\vspace{-1mm}
In this work, we investigated the diversity and bias of synthetic data and synthetic-based FR models. As synthetic data and synthetic models are becoming established as a real alternative to authentic data, it is highly necessity to investigate this in more detail to study discriminatory behavior and to reduce it in future work. In our investigation, we analyzed the distribution and intra-identity attribute consistency on three demographic (gender, ethnicity, age) and one non-demographic (head pose) attribute. The results show that the generator models tend to recreate the distribution from the training datasets and might slightly amplify the imbalance in the synthetic datasets. To investigate the bias, and performance differences depending on different subsets, we performed several experiments on gender, ethnicity, age, and pose splits. The results show, that similar biases can be observed in synthetic FR models as in authentic FR models, motivating existing and new works regarding bias mitigation also be applied to the novel synthetic-based models, especially given the possibility of inducing variability in the synthesized images. {\color{black} Furthermore, our investigation showed that the synthetic face recognition models yet do not achieve the same performance as models trained on authentic data.}
\vspace{-2mm}
\paragraph{Acknowledgement}
\small
This research work has been funded by the German Federal Ministry
of Education and Research and the Hessian Ministry of Higher Education,
Research, Science, and the Arts within their joint support of the National
Research Center for Applied Cybersecurity ATHENE.

%%%%%%%%% REFERENCES
{\small
\bibliographystyle{ieee_fullname}
\bibliography{PaperForReview}
}

\end{document}